\pdfoutput=1

\documentclass[11pt]{article}

\usepackage[]{acl}
\usepackage{times}
\usepackage{latexsym}

\usepackage[T1]{fontenc}

\usepackage[utf8]{inputenc}

\usepackage{microtype}

\usepackage{graphicx}
\usepackage{multirow}
\usepackage{adjustbox}
\usepackage{booktabs}

\usepackage{caption}
\usepackage{subcaption}

\usepackage{inconsolata}

\definecolor{airforceblue}{rgb}{0.36, 0.54, 0.66}
\definecolor{blue(ncs)}{rgb}{0.0, 0.53, 0.74}
\definecolor{blush}{rgb}{0.87, 0.36, 0.51}
\definecolor{softgreen}{rgb}{0.7, 0.9, 0.7}
\definecolor{darkgreen}{rgb}{0.0, 0.5, 0.0} 
\definecolor{darkred}{rgb}{0.5, 0.0, 0.0} 

\title{Resilience of Large Language Models for Noisy Instructions}

\author{
Bin Wang\textsuperscript{$\diamondsuit$}, 
Chengwei Wei\textsuperscript{$\diamondsuit$}, 
Zhengyuan Liu\textsuperscript{$\diamondsuit$}, 
Geyu Lin\textsuperscript{$\diamondsuit$}, 
Nancy F. Chen\textsuperscript{$\diamondsuit,\dag$}
\\
\textsuperscript{$\diamondsuit$}Institute for Infocomm Research (I$^2$R), A*STAR, Singapore\\
\textsuperscript{$\dag$}Centre for Frontier AI Research (CFAR), A*STAR, Singapore\\
\texttt{wang\_bin@i2r.a-star.edu.sg} \\
}

\begin{document}
\maketitle

\begin{abstract}

    As the rapidly advancing domain of natural language processing (NLP), large language models (LLMs) have emerged as powerful tools for interpreting human commands and generating text across various tasks. Nonetheless, the resilience of LLMs to handle text containing inherent errors, stemming from human interactions and collaborative systems, has not been thoroughly explored. Our study investigates the resilience of LLMs against five common types of disruptions including 1) ASR (Automatic Speech Recognition) errors, 2) OCR (Optical Character Recognition) errors, 3) grammatical mistakes, 4) typographical errors, and 5) distractive content. We aim to investigate how these models react by deliberately embedding these errors into instructions. Our findings reveal that while some LLMs show a degree of resistance to certain types of noise, their overall performance significantly suffers. This emphasizes the importance of further investigation into enhancing model resilience. In response to the observed decline in performance, our study also evaluates a "re-pass" strategy, designed to purify the instructions of noise before the LLMs process them. Our analysis indicates that correcting noisy instructions, particularly for open-source LLMs, presents significant challenges.

\end{abstract}

\section{Introduction}

    Large language models offer unprecedented capabilities in understanding and generating human-like text~\cite{touvron2023llama,tunstall2023zephyr}. Built upon the foundation of pre-trained language models (PLMs)~\cite{wei2023overview}, large language models inherit and significantly extend the capabilities of their predecessors by following human-readable instructions, enabling a broad spectrum of applications that were previously challenging or infeasible with unavailable training samples~\cite{kojima2022large}. 

    %%%%%%%%%%%%%%%%%%%%%%%%%%%%%%%%%%%%%%%%%%%%%%%
    \begin{figure}[t]
        \centering
         \includegraphics[width=0.46\textwidth]{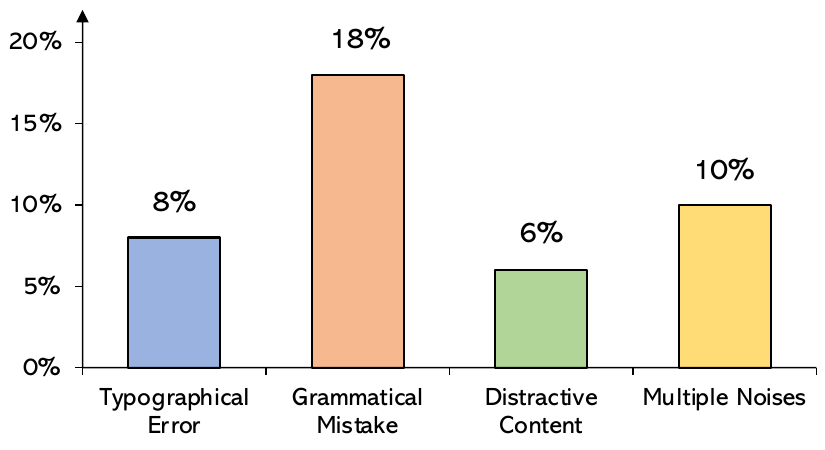}
        \caption{Our analysis scrutinized 500 inputs from real users, focusing on three distinct types of noise. The findings reveal that more than 40\% of the inputs to the model are affected by noise.}
        \label{fig:sharegpt_analysis}
    \vspace{-0.2cm}
    \end{figure}
    %%%%%%%%%%%%%%%%%%%%%%%%%%%%%%%%%%%%%%%%%%%%%%%
    
    Meantime, the capability of LLMs to process noisy instructions is a critical feature that enables their applications in real-world scenarios, where data contains imperfections. To validate the extent of such occurrences, we analyzed the noise within user instructions to a chatbot. Specifically, these instructions were evaluated using GPT-4~\cite{achiam2023gpt} to detect the presence of the specific noise types. Our statistical analysis, illustrated in Figure~\ref{fig:sharegpt_analysis}, indicates that over 40\% of user inputs contain typographical errors, grammatical mistakes, or unrelated content in addition to their primary query\footnote{The dataset for this study comprises user inputs sourced from the ShareGPT dataset, as referenced in~\citet{vicuna2023}.}. Previous research also reveals that human users are inclined to commit errors when interacting with chatbot~\cite{james2013errors}. It is also treated as an evident social cue for human communications~\cite{buhrke2021making}. Therefore, examining LLM's proficiency in managing noisy text inputs is critical to practical applications.

    In our study on deploying Large Language Models (LLMs) across various applications, we categorized the kinds of noisy instructions from three primary sources. First, from a linguistic standpoint, our focus is on grammatical mistakes and typographical errors. Second, we explore noise stemming from system integration, specifically errors originating from Optical Character Recognition (OCR) and Automatic Speech Recognition (ASR) technologies. Lastly, we investigate the impact of destructive content from previous interactions or extended contexts. This part of our study aims to assess the models' proficiency in isolating current queries from past interactions, evaluating their effectiveness in disregarding irrelevant content.
    
    We observe distinct performance impacts across three open and closed-sourced models, when faced with different noise types. First, we find that a higher resilience of models to grammatical mistakes, likely because these errors are also present in data used for pre-training and supervised fine-tuning as also revealed in Figure~\ref{fig:sharegpt_analysis}. This familiarity enables models to more accurately interpret the intended meaning despite such inaccuracies. In contrast, errors from ASR and OCR systems, which are less common in training datasets, present more significant challenges for the models. Furthermore, our study highlights that models are susceptible to being influenced by previous instructions in both cooperative and non-cooperative manner, which can lead to deviations in responses in subsequent interactions. This suggests a limitation in the models' ability to filter out irrelevant or distracting content from past exchanges.
    
    As noisy instructions can be harmful to model perfomrnace, we investigate the potential of leveraging LLMs to mitigate the impact of noisy instructions through a "re-pass" strategy. This approach involves a two-step process: initially, we employ an LLM to conduct zero-shot text normalization to purify the noisy instructions. Next, we prompt the model to process upon the cleaned instruction. Our findings reveal that not all models are adept at fulfilling this role of data normalization. The exception is ChatGPT, which demonstrates a comprehensive understanding of the text and can recover the instruction with different types of noises.

\section{Related Work}

    The progression of general-purpose Large Language Models (LLMs) such as ChatGPT~\cite{achiam2023gpt}, Gemini~\cite{team2023gemini}, LLaMa~\cite{touvron2023llama}, Mistral~\cite{jiang2023mistral}, and Gemma~\cite{gemma} has facilitated a myriad of real-world applications. These advancements are attributed to their capabilities in managing long-range textual dependencies, enhancing contextual comprehension, and displaying a remarkable ability to adapt to a wide array of tasks with minimal need for detailed, task-specific training. Meantime, several recent studies have demonstrated that the user prompt significantly influences task performance, highlighting its indispensable role in the process~\cite{wang-etal-2024-seaeval,zhu2023promptbench}. The following will introduce studies on prompt sensitivity and noisy text reconstruction as related work.

    \begin{figure*}
         \centering
             \includegraphics[width=1.00\textwidth]{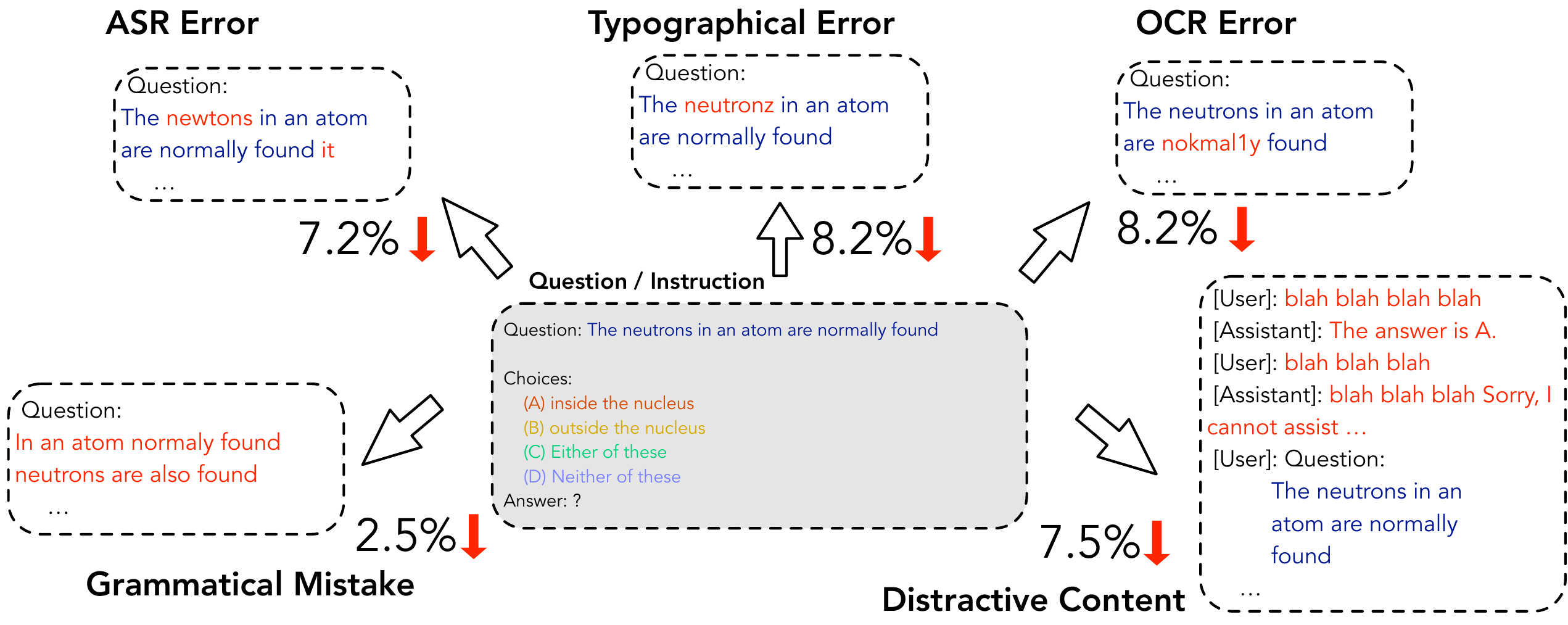}
            \caption{
            Our study identifies and assesses the impact of five distinct categories of textual disruptions on the ChatGPT-3.5 model's effectiveness. We noted a reduction in accuracy between 2.5\% to 8.2\% across the MMLU dataset, a phenomenon directly linked to these varied types of noisy instructions.
            }
            \label{fig:framework}
    \end{figure*}

    \subsection{Instruction Sensitivity}

        Pre-trained large language models exhibit performance variability even with semantically similar inputs. SeaEval~\cite{wang-etal-2024-seaeval} demonstrates that across five different input templates, performance can fluctuate between 5\% and 10\%, depending on the dataset. Similar observations are reported in~\citet{sclar2023quantifying}, highlighting this instability. Moreover, introducing brief sentences such as "Let's think step by step" can significantly enhance performance on reasoning tasks, further underscoring the LLMs' sensitivity~\cite{kojima2022large}. Leveraging this characteristic, various studies concentrate on decomposing and crafting improved prompts to efficiently tackle tasks. \citet{zhou2022large} proposes an additional LLM as a prompt engineer to automatically create prompts that enhance performance. Alternatively, some approaches advocate for the use of search~\cite{prasad-etal-2023-grips} or optimization techniques~\cite{khot2022decomposed,hao2022optimizing,prasad-etal-2023-grips} to identify superior instructions, replacing those that are less effective.

        As an extensive benchmark for adversarial prompts, PromptBench~\cite{zhu2023promptbench} offers a platform to evaluate the resilience of LLMs against attacks across four levels: character, word, sentence, and semantic. In contrast to their research which uses models like DeepWordBug~\cite{gao2018black} and TextBugger~\cite{li2018textbugger}, our work does not aim to introduce adversarial prompts to mislead the model into making errors. Instead, we seek to replicate real-use scenarios where errors might naturally occur and be harmless. Additionally, we categorize different instruction noises, offering a more comprehensive analysis.

    \subsection{Reconstruction of Noisy Instructions}

        Our study extends beyond merely assessing the model's resilience to textual noise; it delves into deploying a comprehensive model designed to rectify a wide spectrum of input errors. The literature review uncovers a variety of approaches specifically devised to address the multifaceted errors highlighted in our research, each tailored to its unique context. These methodologies span from Automatic Speech Recognition (ASR) Error Correction~\cite{mani2020asr,leng2021fastcorrect,jiang2023speech}, which aims to amend errors in speech-to-text transcriptions, to Grammar Error Correction~\cite{yuan-briscoe-2016-grammatical,bryant2023grammatical}, focusing on rectifying grammatical inaccuracies in written text. Additionally, Typographical Error Correction~\cite{church1991probability,zhang-etal-2020-spelling} methods are explored to fix misspellings and typographical mistakes, while Optical Character Recognition (OCR) Error Correction~\cite{tong-evans-1996-statistical,soper-etal-2021-bart,nguyen2021survey} seeks to correct errors introduced during the digitization of printed texts.

\section{Noisy Instruction and Analysis}

    In this section, we describe our methodology for integrating five types of noise into the MMLU benchmark~\cite{hendryckstest2021}, where the original text is devoid of any noise. We employ hybrid rule-based techniques to introduce noise for OCR and Typographical errors. For ASR and Grammatical errors, we leverage a language model to capture error patterns and simulate these errors through a generative process. To simulate distractive content, we embed actual dialogues as irrelevant background information. An illustration of each type of noisy dataset is provided in Figure~\ref{fig:framework} and a summarization is shown in Table~\ref{tab:noisytext}.

        \begin{figure*}
             \centering
                 \centering
                 \includegraphics[width=\textwidth]{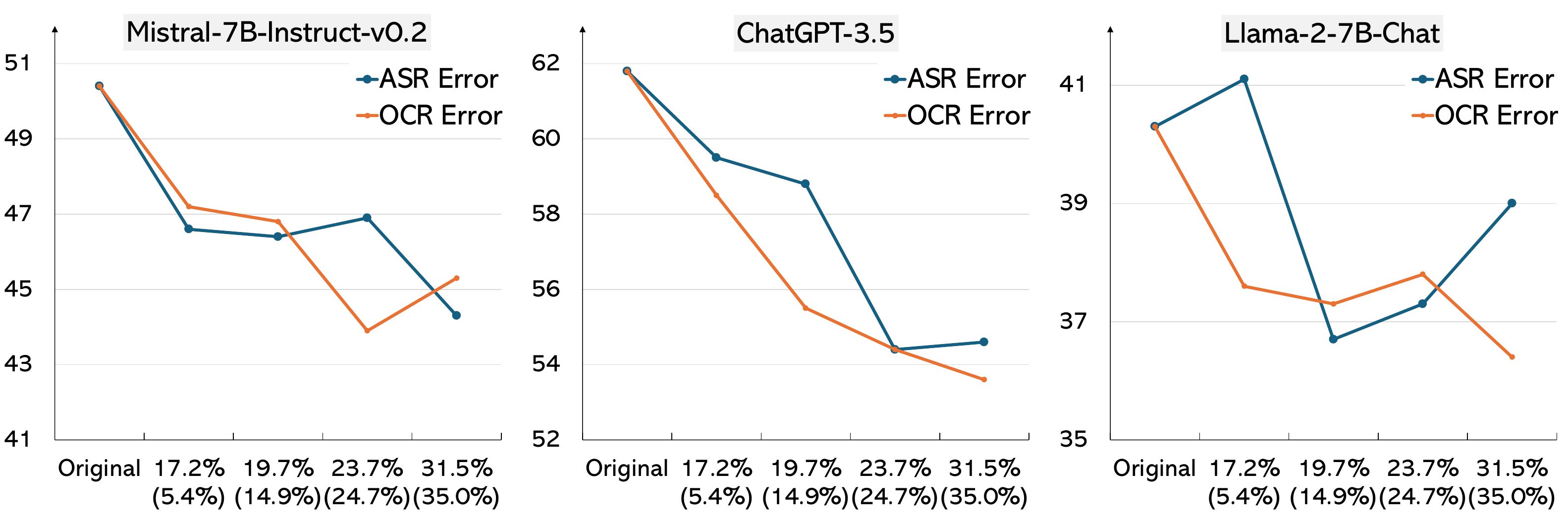}
                \caption{
                Evaluation of the performance of three Large Language Models (LLMs) using the adapted MMLU dataset, emphasizing different error ratios, as measured by Word Error Rate (WER). The x-axis represents the WER values for Automatic Speech Recognition (ASR) and Optical Character Recognition (OCR), indicated within brackets. The performance declines with noisy instructions.
                }
                \label{fig:2-1}
        \end{figure*}

        %%%%%%%%%%%%%%%%%%%%%%%%%%%%%%%%%%%%%%%%%%%%%%%
        \begin{table}[t]
            \centering
            \begin{adjustbox}{width=0.5\textwidth,center}
            \begin{tabular}{ c | c }
            \toprule
             \textbf{Noise Type} & \textbf{Sources} \\ \hline
             \multirow{2}{*}{ASR} & LibriSpeech~\cite{panayotov2015librispeech} \\ 
              & CommonVoice-15~\cite{ardila-etal-2020-common} \\ \hline
             \multirow{2}{*}{OCR} & NLPAug~\cite{ma2019nlpaug} \\ 
              & OCR Engine \\ \hline
             \multirow{2}{*}{Grammatical} & JELEG~\cite{napoles-etal-2017-jfleg} \\ 
              & C4-200M~\cite{stahlberg-kumar-2021-synthetic} \\ \hline
             \multirow{2}{*}{Typographical} & NLPAug~\cite{ma2019nlpaug} \\ 
              & Keyboard, Spelling, Random \\ \hline
             Distractive Content & ShareGPT~\cite{vicuna2023} \\ 
            \bottomrule
            \end{tabular}
            \end{adjustbox}
            \caption{
                A summary of the techniques and datasets.
            }
            \label{tab:noisytext}
        \end{table}
        %%%%%%%%%%%%%%%%%%%%%%%%%%%%%%%%%%%%%%%%%%%%%%%

    \subsection{Automatic Speech Recognition (ASR)}

        \subsubsection{Method}

            Given that the original texts are not available in audio format, we propose employing a generative model to effectively replicate the patterns of ASR errors. This method enables us to inject realistic ASR errors into pre-existing textual materials. Specifically, we utilize one of the premier ASR models, Whisper-Tiny~\cite{radford2023robust}, as our ASR engine ($\mathbf{M}_{ASR}$). We utilize the CommonVoice-15~\cite{ardila-etal-2020-common} dataset and the noisy test set from the LibriSpeech~\cite{panayotov2015librispeech} dataset as the source data. These datasets together offer over 1,000 hours of speech data, all of which are processed by our ASR engine without prior exposure (zero-shot). We then generate the ASR output texts ($T_n$), which include ASR-induced errors, by processing the simulated audio through our ASR engine $T_n=M_{ASR}(T_c)$, resulting in outputs that diverge from the original clean transcripts ($T_c$).
    
            Using the provided dataset of clean and predicted transcripts, we divide them into four distinct categories based on their Word-Error-Rate (WER): less than 10\%, 10\%-20\%, 20\%-30\%, and 30\%-40\%, allocating 80,000 samples to each category. By leveraging this paired data, we finetune Tiny-Llama-1.1B-Chat~\cite{zhang2024tinyllama} models for each category to learn the underlying patterns of ASR errors.
            We use the trained error generation model to introduce ASR errors into each question sentence from the MMLU dataset independently. This approach produces a varied WER (Word Error Rate), as illustrated in Figure~\ref{fig:2-1}. This noise injection method is consistently applied across all types of noise, except for the destructive content. The same set of 1,000 questions is selected for five noisy instructions to benchmark the performance.

        \subsubsection{Discussion}

            Figure~\ref{fig:2-1} presents the noisy instructions alongside the corresponding accuracy. ASR noise embedded in the instructions is harmful to all models. As the WER increases, the magnitude of performance accuracy drops accordingly. This trend highlights a critical vulnerability in current models when dealing with speech recognition errors. It is worth noticing that the close-sourced ChatGPT-3.5 model is as vulnerable as open-sourced models like Mistral and Llama.

            Although LLMs have numerous applications in processing spoken content, they lack robustness against errors introduced by ASR systems. Consequently, there is a critical need for the development of LLMs that are resilient to ASR errors, as well as the creation of comprehensive speech-to-text foundation models that can directly handle speech inputs~\cite{Qwen-Audio,tang2024salmonn,wang2024audiobench,zhang2024mowe}.

        \begin{figure*}
             \centering
                 \centering
                 \includegraphics[width=\textwidth]{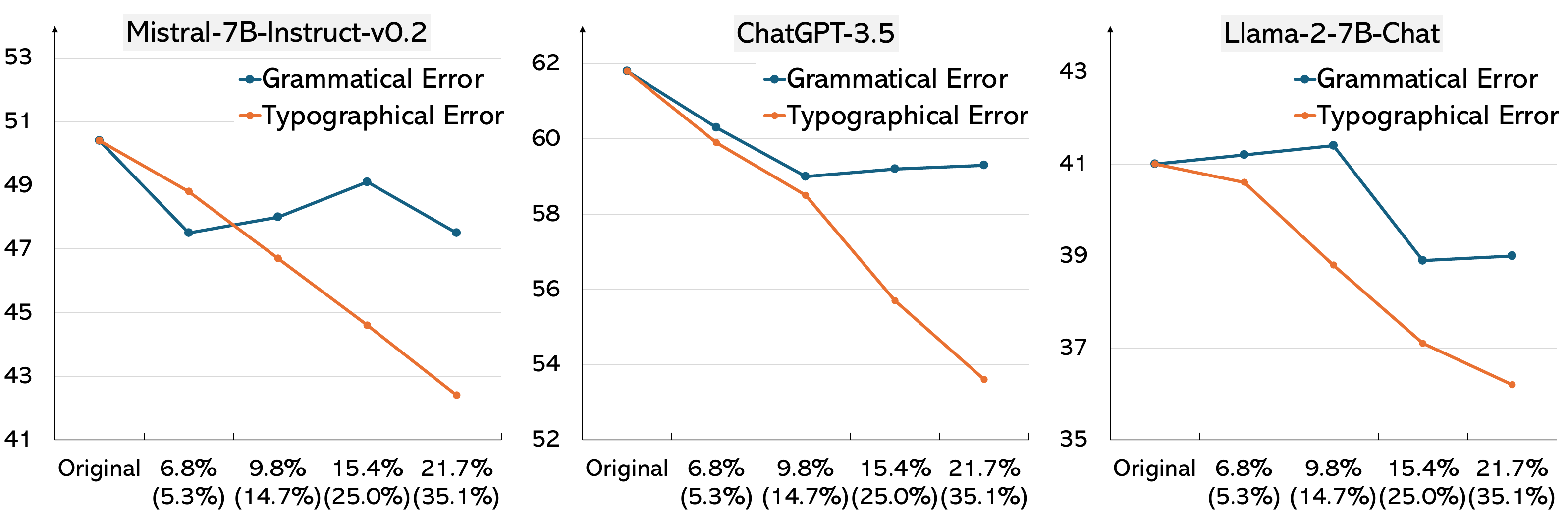}
                \caption{Evaluation of the performance of three Large Language Models (LLMs). The x-axis represents the WER values for grammatical mistakes and typographical errors, indicated within brackets.}
                \label{fig:3}
        \end{figure*}

    \subsection{Optical Character Recognition (OCR)}

        \subsubsection{Method}

            OCR technology is prone to specific types of errors, often misclassifying items that appear visually similar, especially on word-level. To simulate OCR errors, we employed the OcrAug engine~\cite{ma2019nlpaug}, enhancing it with broader OCR mapping dictionaries to inject errors into clean text. 
            We expanded the initial OCR error dictionary from 12 to 36 groups of characters prone to misclassification. Each word is altered by replacing it with versions that include easily misclassified characters, introducing OCR errors with variations of 1 to 3 characters. To simulate varying degrees of OCR error severity, we adjust the number of words altered, categorizing them into four distinct groups following the above convention.

        \subsubsection{Discussion}

            The findings, in Figure~\ref{fig:2-1}, reveal that all models demonstrate a lack of robustness to OCR errors, showing a higher performance decline compared to similar WER from ASR errors. This may stem from the characteristics of LLMs. First, LLMs use BPE tokens~\cite{gage1994new} for pre-processing, meaning the corrupted words will not follow the original tokenization scheme. Such discrepancies can result in words being split into multiple tokens, significantly disrupting the original semantic representation. In contrast, ASR errors tend to perverse word integrity. Additionally, the pre-training phase for these models seldom includes text with OCR-induced errors, which are rooted in visual effects. Enhancing OCR robustness of LLMs should include both pre-training exposure and tokenization strategies. However, it is important to note that character-level tokenization~\cite{xue2022byt5}, despite its potential benefits, is still inferior to common subword tokenization methods.

    \subsection{Grammatical Mistakes}

        \subsubsection{Method}

            To replicate grammatical errors, similar to those produced for ASR systems, we conducted training on a generative model to emulate this pattern. Our approach involves utilizing two primary sources to gather pairs for both clean and noisy text containing grammatical errors: JELEG~\cite{napoles-etal-2017-jfleg} and C4-200M~\cite{stahlberg-kumar-2021-synthetic}. Both datasets serve the purpose of grammatical error corrections, and we employ their pairs in reverse sequence, thereby transitioning from grammatical error correction to error injection. Four models are trained to learn error patterns with four distinct WER ranges and subsequently applied to each question sentence from the MMLU dataset to simulate grammatical mistakes.

        \subsubsection{Discussion}

            The performance on the MMLU dataset with grammatical errors is shown in Figure~\ref{fig:3}. We observe that LLMs exhibit a more resilient performance in handling grammatical mistakes. Specifically, the performance deterioration of LLM when dealing with grammatical errors is less severe compared with other types of errors. This suggests that LLMs possess a certain degree of robustness to grammatical mistakes, indicating their ability to process contextualized information even with grammatical deficiencies. We expect that the LLM pre-training and fine-tuning stages have been exposed to a reasonable amount of content with grammatical mistakes, which aligns with our findings shown in Figure~\ref{fig:sharegpt_analysis}.

    \subsection{Typographical Errors}

        \subsubsection{Method}

            To address typographical errors, we utilize a hybrid approach that combines three character-level modifications, as implemented in the NLPAug package~\cite{ma2019nlpaug}.
            The modifications are derived from three primary sources to construct text with typographical errors: 1) Spelling errors, comprising 13,000+ groups of commonly misspelled words. 2) Keyboard errors, which simulate errors arising from mistyping characters that are physically close to each other on the keyboard. 3) Random errors, where characters are arbitrarily replaced by others. Each type of error is equally likely to occur, ensuring a diverse representation of typographical errors in the generated noisy text. The word error rate is improved by adjusting the number of words that are altered and each word can have a maximum of 3 altered characters for spelling and random errors. The number of adapted words is changed to be categorized into four distinct WER groups.

        \subsubsection{Discussion}

            Figure~\ref{fig:3} presents the results with the WER specified in parentheses for typographical errors. From the results, we can see that the performance of LLM is severely influenced by typographical errors. The analysis in Figure~\ref{fig:sharegpt_analysis} reveals that a fraction of text data contains typographical errors. These errors often result in tokenization issues similar to those observed with OCR errors, contributing to the reduction in performance. 
        
    \begin{figure}[t]
         \centering
            \includegraphics[width=0.5\textwidth]{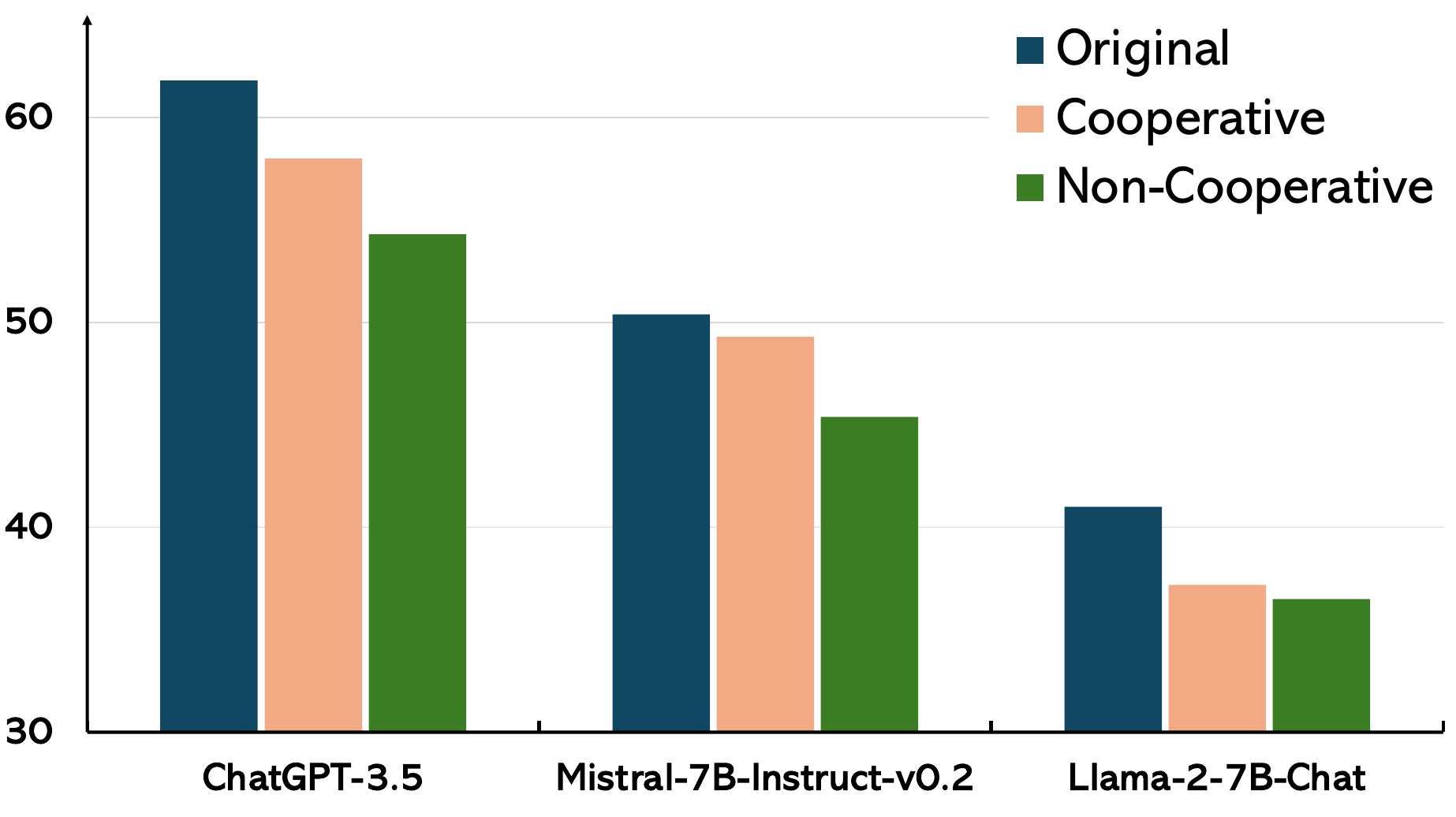}
            \caption{The performance of LLMs with both cooperative and non-cooperative distactive content. Both lead to performance declines while non-cooperative distractions have a more disruptive impact.}
            \label{fig:distractive}
    \end{figure}

    \subsection{Distractive Content}

        \subsubsection{Methods}

            When interacting with large language models, users may introduce irrelevant information into their input for multifaceted reasons. This can occur due to a lack of clarity about previous interactions, retrieval of unrelated documents in Retrieval-Augmented Generation (RAG) systems, or simply by accident. With historical content, LLM can grasp the context more effectively and deliver responses that are more tailored to the context. However, the impact of irrelevant content on the performance related to the most recent instructions remains uncertain. Therefore, we study the effect of irrelevant content on its influence on the current instruction. Specifically, we add one turn of irrelevant dialogue content sampled from the ShareGPT dataset~\cite{vicuna2023}, which consists of real-user interactions with other LLMs. The utterance speaker information is injected as shown in Figure~\ref{fig:framework}. Here we study two scenarios of user interactions.

            \noindent\textbf{Cooperative distraction} indicates that the user follows multi-turn dialogue patterns provided by respective models. It can be viewed that the user forgot to clean chat histories while initiating requests, which occurs frequently in human-chatbot interactions. The model must possess the ability to discern the lack of relevance between current instructions and historical information. This capability is essential for ensuring the model does not mistakenly integrate past interactions into current responses, leading to inaccurate responses.

            \noindent\textbf{Non-cooperative distraction} indicates that the irrelevant content is concatenated directly with the current inquiries without following the designed template of the particular chatbot model. In instances where RAG systems are employed, retrieving content irrelevant to the current instruction is possible. When concatenated with the instruction, such unrelated content can adversely affect the responses.

        \subsubsection{Discussion}

            Figure~\ref{fig:distractive} demonstrates the impact of both cooperative and non-cooperative distractive content on the performance of three models. It reveals that introducing distractions can result in performance decline across all models. As expected, the non-cooperative distractive content exhibits a more significant impact. More specifically, the analysis indicates a performance decline of \emph{ChatGPT-3.5} by 3.8\% and 7.5\% for cooperative and non-cooperative distractions, respectively. This trend is consistent across other models such as \emph{Mistral-7B-Instruct-v0.2} and \emph{Llama-2-7B-Chat}. 
            
            In cooperative settings, this decline indicates the models' inability to completely disregard irrelevant dialogue history while processing current requests, as responses tend to be context-dependent. While the capability to generate context-dependent responses based on dialogue history can be advantageous, our findings suggest that it becomes harmful when the history consists of irrelevant distractions. This may be because the models are commonly tuned for multi-turn dialogue instructions, where context dependency is emphasized and irrelevant context is rarely introduced. Therefore, enhancing models' ability to discern relevant from irrelevant content is crucial in further model development to show higher robustness in handling distractions. On the other hand, non-cooperative settings present even higher challenges for the model in terms of isolating irrelevant content. It is particularly crucial for systems augmented with Retrieval-Augmented Generation (RAG), where retrieving irrelevant information from the database can lead to performance decline. Consequently, dynamic retrieval strategies and filtering techniques are necessary to enhance the robustness of models towards distractions and maintain optimal functionality as discussed in~\citet{asai2023self}.

    %%%%%%%%%%%%%%%%%%%%%%%%%%%%%%%%%%%%%%%%%%%%%%%
    \begin{table*}[t]
        \centering
        \begin{adjustbox}{width=0.95\textwidth,center}
        \begin{tabular}{ c | c | c | c | c | c }
        \toprule
        \textbf{Harmonizer} & \textbf{WER} & \textbf{Base Acc} & \textbf{ChatGPT-3.5} & \textbf{Mistral-7B-Instruct-v0.2} & \textbf{Llama-2-7B-Chat} \\  \hline
        \hline
        Clean & 0\% & 50.4\% & \textcolor{darkgreen}{+0.4\%} (50.8\%) & \textcolor{darkred}{-3.8\%} (46.6\%) & \textcolor{darkred}{-5.3\%} (45.1\%) \\ 
        \hline \hline
        \multirow{4}{*}{ASR Error} & 17.2\% & 46.6\% & \textcolor{darkgreen}{+2.9\%} (49.5\%) & \textcolor{darkgreen}{-1.4\%} (48.0\%) &  \textcolor{darkred}{-1.2\%} (45.4\%) \\ 
         & 19.7\% & 46.4\% & \textcolor{darkgreen}{+3.3\%} (49.7\%) & \textcolor{darkgreen}{+0.8\%} (47.2\%) & \textcolor{darkred}{-0.8\%} (45.6\%) \\ 
         & 23.7\% & 46.9\% & \textcolor{darkgreen}{+3.2\%} (50.1\%) & \textcolor{darkgreen}{+1.9\%} (48.8\%) & \textcolor{darkred}{-2.2\%} (44.7\%) \\ 
         & 31.5\% & 44.3\% & \textcolor{darkgreen}{+2.3\%} (46.6\%) & \textcolor{darkgreen}{+2.2\%} (46.5\%) & \textcolor{darkred}{-1.6\%} (42.7\%) \\ 
        \hline

        \multirow{4}{*}{OCR Error} & 5.4\% & 47.2\% & \textcolor{darkgreen}{+3.1\%} (50.3\%) & \textcolor{darkred}{-0.1\%} (47.1\%) & \textcolor{darkred}{-0.6\%} (46.6\%) \\ 
         & 14.9\% & 46.8\% & \textcolor{darkgreen}{+1.7\%} (48.5\%) & \textcolor{darkgreen}{+0.9\%} (47.7\%) & \textcolor{darkred}{-0.7\%} (46.1\%) \\ 
         & 24.7\% & 43.9\% & \textcolor{darkgreen}{+5.0\%} (48.9\%) & \textcolor{darkgreen}{+3.0\%} (46.9\%) & \textcolor{darkgreen}{+0.3\%} (44.2\%) \\ 
         & 35.0\% & 45.3\% & \textcolor{darkgreen}{+0.6\%} (45.9\%) & \textcolor{darkgreen}{+1.7\%} (47.0\%) & \textcolor{darkred}{-3.3\%} (42.0\%) \\ 
        \hline

        \multirow{4}{*}{Grammatical Error} & 6.8\% & 47.5\% & \textcolor{darkgreen}{+2.3\%} (49.8\%) & \textcolor{darkgreen}{+2.4\%} (49.9\%) & \textcolor{darkred}{-3.0\%} (44.5\%) \\ 
         & 9.8\% & 48.0\% & \textcolor{darkgreen}{+1.7\%} (49.7\%) & \textcolor{darkgreen}{+0.4\%} (48.4\%)  & \textcolor{darkred}{-2.1\%} (45.9\%) \\ 
         & 15.4\% & 49.1\% & \textcolor{darkgreen}{+1.2\%} (50.3\%) & \textcolor{darkred}{-1.6\%} (47.5\%) & \textcolor{darkred}{-5.0\%} (44.1\%) \\ 
         & 21.7\% & 47.5\% & \textcolor{darkgreen}{+2.2\%} (49.7\%) & \textcolor{darkgreen}{+0.1\%} (47.6\%) & \textcolor{darkred}{-4.6\%} (42.9\%) \\ 
        \hline

        \multirow{4}{*}{Typographical Error} & 5.3\% & 48.8\% & \textcolor{darkgreen}{+1.3\%} (50.1\%) & \textcolor{darkgreen}{+0.5\%} (49.3\%) & \textcolor{darkred}{-2.6\%} (46.2\%) \\ 
         & 14.7\% & 46.7\% & \textcolor{darkgreen}{+3.0\%} (49.7\%) & \textcolor{darkgreen}{+1.0\%} (47.7\%)  & \textcolor{darkred}{-2.5\%} (44.2\%) \\ 
         & 25.0\% & 44.6\% & \textcolor{darkgreen}{+6.4\%} (51.0\%) & \textcolor{darkgreen}{+3.0\%} (47.6\%)  & \textcolor{darkred}{-0.5\%} (44.1\%) \\ 
         & 35.1\% & 42.4\% & \textcolor{darkgreen}{+5.0\%} (47.4\%) & \textcolor{darkgreen}{+1.9\%} (44.3\%)  & \textcolor{darkred}{-0.1\%} (42.3\%) \\ 
        
        \bottomrule
        \end{tabular}
        \end{adjustbox}
        \caption{Performance evaluation of \emph{Mistral-7B-Instruct-v0.2} on modified noisy MMLU datasets corrected by three different Large Language Models (LLMs). "WER" represents the word-error-rate and "Base Acc" refers to the initial accuracy of noisy dataset prior to any corrections applied using LLMs.} 
        \label{tab:harmonizer}
    \end{table*}
    %%%%%%%%%%%%%%%%%%%%%%%%%%%%%%%%%%%%%%%%%%%%%%%

\section{Recovery of Noisy Instructions}

        Previous research has demonstrated the capability of language models to amend specific errors~\cite{ma2023can,mai2024enhancing}. In this section, we explore the effectiveness of utilizing Large Language Models (LLMs) for zero-shot correction of four previously identified types of noise.

    \subsection{Methods}

\begin{figure}[t]
         \centering
            \includegraphics[width=0.33\textwidth]{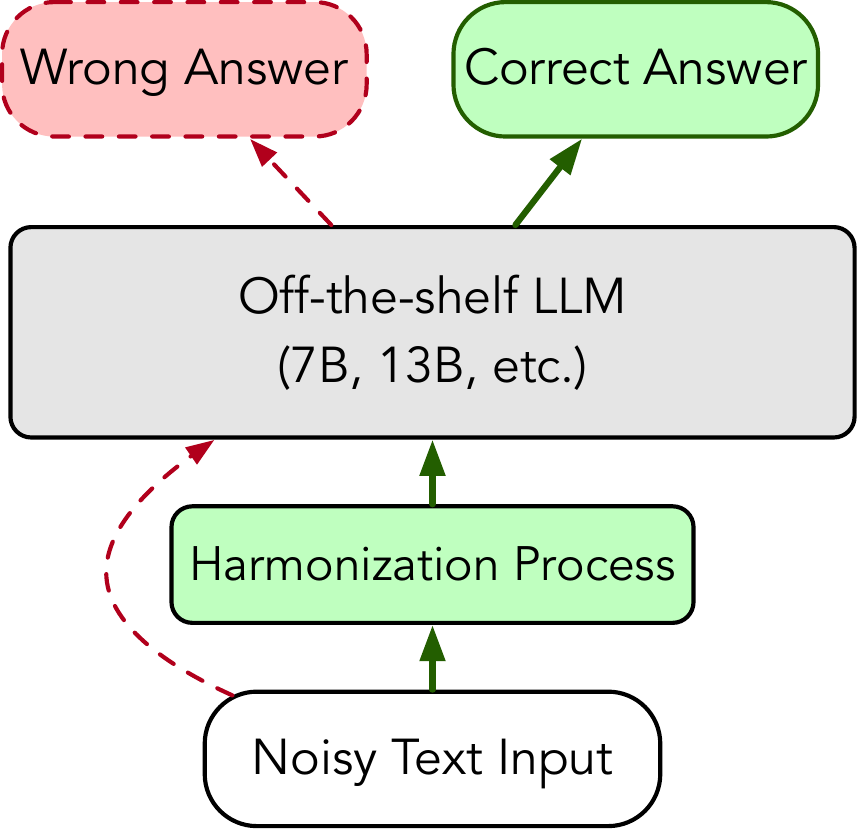}
            \caption{
            The "re-pass" strategy involves a preliminary step where noisy instructions undergo a harmonization process to refine before input into the model for generating responses.}
            \label{fig:framework_harmonizer}
    \end{figure}

        We employ a "re-pass" strategy to investigate whether LLMs can be used to recover clean instruction from noisy counterparts. As shown in Figure~\ref{fig:framework_harmonizer}, the noisy instructions are processed with a large language model (e.g. \emph{ChatGPT-3.5}, \emph{Mistral-7B-Instruct-v0.2}) to correct errors contained in the instructions. After that, the revised instruction is fed into the task-solving LLM to perform the desired task.
        
    %%%%%%%%%%%%%%%%%%%%%%%%%%%%%%%%%%%%%%%
    %%%%%%%%%%%%%%%%%%%%%%%%%%%%%%%%%%%%%%%

    %%%%%%%%%%%%%%%%%%%%%%%%%%%%%%%%%%%%%%%%%%%%%%%
    \begin{table*}[t]
        \centering
        \begin{adjustbox}{width=0.91\textwidth,center}
        \begin{tabular}{ l | l }
        \toprule
         %\textbf{Category} & \textbf{Instruction}  \\  \hline
         \multicolumn{2}{c}{\emph{\textbf{Example \#1}}} \\
        \hline
         \multirow{2}{*}{Clean Instruction} & An American firm moves a manufacturing plant from the United States to Brazil. How \\ 
         & will this affect gross domestic product (GDP) in the United States and in Brazil? \\ \hline
         \multirow{2}{*}{Noisy Instruction (ASR)} & \textcolor{red}{and} american firm \textcolor{red}{moved} manufacturing plant from the united states to brazil. how \\
         & \textcolor{red}{well} this affect gross domestic product \textcolor{red}{g p} in the united states and \textcolor{red}{(void)} brazil \\ \hline
         \multirow{2}{*}{\emph{ChatGPT-3.5}} & An American firm \textcolor{red}{moved} its manufacturing plant from the United States to Brazil. How\\
         & will this affect the gross domestic product (GDP) in the United States and \textcolor{red}{(void)} Brazil? \\ \hline
         \multirow{2}{*}{\emph{Mistral-7B-Instruct-v0.2}} & An American firm \textcolor{red}{moved} its manufacturing plant from the United States to Brazil. How \\
         & will this affect gross domestic product (GDP) in the United States and \textcolor{red}{(void)} Brazil? \\ \hline
         \multirow{2}{*}{\emph{Llama-2-7b-Chat}} & \textcolor{red}{(void)} American firm \textcolor{red}{moved} manufacturing plant from the United States to Brazil. \\
         & \textcolor{red}{(-- void --)}\\

            \midrule
         \multicolumn{2}{c}{\emph{\textbf{Example \#2}}} \\ \hline
         
         \multirow{2}{*}{Clean Instruction} & Darwin's mechanism of natural selection required long time spans in order to modify  \\ 
         & species. From whom did Darwin get the concept of Earth's ancient age? \\ \hline
         
         \multirow{2}{*}{Noisy Instruction (Grammar)} & Darwin's mechanism of natural selection \textcolor{red}{require} long time spans in order to modify \\
         & species. From whom \textcolor{red}{Darwin got the concept of Earth's old age?} \\ \hline
         
         \multirow{2}{*}{\emph{ChatGPT-3.5}} & Darwin's mechanism of natural selection requires long time spans in order to modify \\
         & species. From whom did Darwin get the concept of \textcolor{red}{(void)} Earth's \textcolor{red}{old} age? \\ \hline
         
         \multirow{2}{*}{\emph{Mistral-7B-Instruct-v0.2}} & Darwin's \textcolor{red}{theory} of natural selection requires long time spans to modify species. \\
         & From whom did Darwin get the concept of the Earth's \textcolor{red}{old} age? \\ \hline
         
         \multirow{2}{*}{\emph{Llama-2-7b-Chat}} & Darwin's mechanism of natural selection requires long time spans in order to modify \\
         & species. \textcolor{red}{(-- void --)} \\
         
        \bottomrule
        \end{tabular}
        \end{adjustbox}
        \caption{Two qualitative examples of noisy instruction correction using LLMs.} 
        \label{tab:examples}
    \end{table*}
    %%%%%%%%%%%%%%%%%%%%%%%%%%%%%%%%%%%%%%%%%%%%%%%

    \subsection{Results and Analysis}

        In Table~\ref{tab:harmonizer}, we show the evaluation results of \emph{Mistral-7B-Instruct-v0.2} model with the instructions being cleaned by three LLMs including \emph{ChatGPT-3.5}, \emph{Mistral-7B-Instruct-v0.2} (self-correction) and \emph{Llama-2-7B-Chat}.

        First, we witness that \emph{Llama-2-7B-Chat} does not show a good capability in error correction of noisy instructions. It even leads to performance drop even after the correction process. We witness that the model could not follow instruction as well as other models and the revised instructions can be modified with hallucinations added, which makes the final answer unanswerable. Therefore, even with clean instruction as input, the performance on corrected instructions drops up to 5.3\%.

        Second, in comparison to \emph{Mistral-7B-Instruct-v0.2}, \emph{ChatGPT-3.5} demonstrates superior performance in detecting and amending errors. Utilizing the re-pass strategy enables the recovery of most mistakes, particularly in samples with a Word Error Rate (WER) of up to 30\%. It is anticipated that a WER exceeding 30\% may often result in damage to the model that cannot be easily reversed.

        Third, the self-correction mechanism of \emph{Mistral} exhibits limited effectiveness. While capable of rectifying certain mistakes, it may inadvertently introduce new errors when handling clean instructions, resulting in a performance decline of up to 3.8\%. Consequently, this creates an unavoidable barrier to deploying such models in real-world applications unless in an environment where noise is guaranteed.

        Regarding types of noise, grammatical errors are typically easier to correct. Such errors are critical to comprehension and do not interfere with the tokenization process much. Therefore, they exert minimal impact on overall performance and are the least challenging noise type to be corrected.

    \noindent\textbf{Qualititive Study.}
        Table~\ref{tab:examples} presents two examples comparing the effectiveness of various models. The results demonstrate that while \emph{ChatGPT-3.5} may not fully restore the original instruction, the resulting instructions are more comprehensible. The worst case is \emph{Llama-2-7B-Chat} which often results in additional information loss from the original instructions.

    \subsection{Discussion on Efficacy}

        In this section, we explore how effective Large Language Models (LLMs) are at mitigating the impact of noisy instructions. However, there are two major drawbacks: 1) open-sourced models generally perform poorly in this task and 2) there is an extra computational cost associated with processing requests. Therefore, there is a need for a lightweight model that is task-agnostic for noisy instruction correction. During our research, we explored fine-tuning an LLM with 1.1B model~\cite{zhang2024tinyllama} sizes using synthesized data to recover clean instructions from noisy ones. However, we found it is challenging for the model to grasp the real intent behind instructions and as a result, unable to performance error corrections accurately. This difficulty is attributed to the constraints imposed by the model size. Therefore, efficient correction of instructions require further investigation, which holds significant use cases like defending adversarial attacks and system integration (e.g. ASR and OCR).

\section{Conclusion}

    In this study, we delve into the resilience of LLMs against noise in instructions from human interactions and system integration. This highlights the complex challenge of processing and recovering accurate information from noisy inputs.  Further, we investigate into the "re-pass" strategy and spot the limitations of current open-sourced models in handling noise corrections. Our findings reemphasise that stronger noisy correction and resilience capabilities are required for LLMs, especially for system integration like ASR and OCR, and process various user requests under both cooperative and no-cooperative settings.

\section{Acknowledgement}

    This research is supported by the National Research Foundation, Singapore and Infocomm Media Development Authority, Singapore under its National Large Language Models Funding Initiative. Any opinions, findings and conclusions or recommendations expressed in this material are those of the author(s) and do not reflect the views of National Research Foundation, Singapore and Infocomm Media Development Authority, Singapore.

\section*{Limitations}

    First, injecting real noise patterns into the evaluation process poses a significant limitation. Simulating authentic noise that accurately reflects the varied and complex errors encountered in real-world data is challenging. This difficulty arises because noise can stem from numerous sources, such as human errors, system glitches, or environmental interference. In this study, we leverage real sample with error pairs, enabling LLM to simulate the error pattern as much as possible. The grasped knowledge is then applied to introduce noise in the aspects of Automatic Speech Recognition (ASR) and grammatical errors.  However, it's important to acknowledge that this process may lead to potential information loss.

    Second, our analysis and error types are limited to English benchmarks without extension to multilingual scenarios. The problem becomes more complex as each language has its own uniqueness. Moreover, coding-switching noise introduces further complexities. The assessment of LLM's resilience to noisy instructions in multilingual scenarios is an area needs future explorations.

    Last, in this study, we focus on five types of noise rooted from system integration (ASR, OCR) and user interactions (Typographical, Grammatical, and Distraction Content). While comprehensive within its defined scope, our work does not encompass all possible sources of noise that could affect LLM performance. For instance, semantic ambiguities, stylistic variations or multilingualism, which could significantly impact the interpretation and processing capabilities of LLMs, are not investigated in detail~\cite{lovenia2024seacrowd,wang-etal-2024-craft,lin2024crossin}.

\bibliography{anthology,custom}

\begin{thebibliography}{50}
\expandafter\ifx\csname natexlab\endcsname\relax\def\natexlab#1{#1}\fi

\bibitem[{Achiam et~al.(2023)Achiam, Adler, Agarwal, Ahmad, Akkaya, Aleman,
  Almeida, Altenschmidt, Altman, Anadkat et~al.}]{achiam2023gpt}
Josh Achiam, Steven Adler, Sandhini Agarwal, Lama Ahmad, Ilge Akkaya,
  Florencia~Leoni Aleman, Diogo Almeida, Janko Altenschmidt, Sam Altman,
  Shyamal Anadkat, et~al. 2023.
\newblock Gpt-4 technical report.
\newblock \emph{arXiv preprint arXiv:2303.08774}.

\bibitem[{Ardila et~al.(2020)Ardila, Branson, Davis, Kohler, Meyer, Henretty,
  Morais, Saunders, Tyers, and Weber}]{ardila-etal-2020-common}
Rosana Ardila, Megan Branson, Kelly Davis, Michael Kohler, Josh Meyer, Michael
  Henretty, Reuben Morais, Lindsay Saunders, Francis Tyers, and Gregor Weber.
  2020.
\newblock \href {https://aclanthology.org/2020.lrec-1.520} {Common voice: A
  massively-multilingual speech corpus}.
\newblock In \emph{Proceedings of the Twelfth Language Resources and Evaluation
  Conference}, pages 4218--4222, Marseille, France. European Language Resources
  Association.

\bibitem[{Asai et~al.(2023)Asai, Wu, Wang, Sil, and Hajishirzi}]{asai2023self}
Akari Asai, Zeqiu Wu, Yizhong Wang, Avirup Sil, and Hannaneh Hajishirzi. 2023.
\newblock Self-rag: Learning to retrieve, generate, and critique through
  self-reflection.
\newblock \emph{arXiv preprint arXiv:2310.11511}.

\bibitem[{Bryant et~al.(2023)Bryant, Yuan, Qorib, Cao, Ng, and
  Briscoe}]{bryant2023grammatical}
Christopher Bryant, Zheng Yuan, Muhammad~Reza Qorib, Hannan Cao, Hwee~Tou Ng,
  and Ted Briscoe. 2023.
\newblock Grammatical error correction: A survey of the state of the art.
\newblock \emph{Computational Linguistics}, 49(3):643--701.

\bibitem[{B{\"u}hrke et~al.(2021)B{\"u}hrke, Brendel, Lichtenberg, Greve, and
  Mirbabaie}]{buhrke2021making}
Johannes B{\"u}hrke, Alfred~Benedikt Brendel, Sascha Lichtenberg, Maike Greve,
  and Milad Mirbabaie. 2021.
\newblock Is making mistakes human? on the perception of typing errors in
  chatbot communication.
\newblock \emph{Proceedings of the 54th Hawaii International Conference on
  System Sciences}.

\bibitem[{Chiang et~al.(2023)Chiang, Li, Lin, Sheng, Wu, Zhang, Zheng, Zhuang,
  Zhuang, Gonzalez, Stoica, and Xing}]{vicuna2023}
Wei-Lin Chiang, Zhuohan Li, Zi~Lin, Ying Sheng, Zhanghao Wu, Hao Zhang, Lianmin
  Zheng, Siyuan Zhuang, Yonghao Zhuang, Joseph~E. Gonzalez, Ion Stoica, and
  Eric~P. Xing. 2023.
\newblock \href {https://lmsys.org/blog/2023-03-30-vicuna/} {Vicuna: An
  open-source chatbot impressing gpt-4 with 90\%* chatgpt quality}.

\bibitem[{Chu et~al.(2023)Chu, Xu, Zhou, Yang, Zhang, Yan, Zhou, and
  Zhou}]{Qwen-Audio}
Yunfei Chu, Jin Xu, Xiaohuan Zhou, Qian Yang, Shiliang Zhang, Zhijie Yan, Chang
  Zhou, and Jingren Zhou. 2023.
\newblock Qwen-audio: Advancing universal audio understanding via unified
  large-scale audio-language models.
\newblock \emph{arXiv preprint arXiv:2311.07919}.

\bibitem[{Church and Gale(1991)}]{church1991probability}
Kenneth~W Church and William~A Gale. 1991.
\newblock Probability scoring for spelling correction.
\newblock \emph{Statistics and Computing}, 1:93--103.

\bibitem[{Gage(1994)}]{gage1994new}
Philip Gage. 1994.
\newblock A new algorithm for data compression.
\newblock \emph{The C Users Journal}, 12(2):23--38.

\bibitem[{Gao et~al.(2018)Gao, Lanchantin, Soffa, and Qi}]{gao2018black}
Ji~Gao, Jack Lanchantin, Mary~Lou Soffa, and Yanjun Qi. 2018.
\newblock Black-box generation of adversarial text sequences to evade deep
  learning classifiers.
\newblock In \emph{2018 IEEE Security and Privacy Workshops (SPW)}, pages
  50--56. IEEE.

\bibitem[{Hao et~al.(2022)Hao, Chi, Dong, and Wei}]{hao2022optimizing}
Yaru Hao, Zewen Chi, Li~Dong, and Furu Wei. 2022.
\newblock Optimizing prompts for text-to-image generation.
\newblock \emph{arXiv preprint arXiv:2212.09611}.

\bibitem[{Hendrycks et~al.(2021)Hendrycks, Burns, Basart, Zou, Mazeika, Song,
  and Steinhardt}]{hendryckstest2021}
Dan Hendrycks, Collin Burns, Steven Basart, Andy Zou, Mantas Mazeika, Dawn
  Song, and Jacob Steinhardt. 2021.
\newblock Measuring massive multitask language understanding.
\newblock \emph{Proceedings of the International Conference on Learning
  Representations (ICLR)}.

\bibitem[{James(2013)}]{james2013errors}
Carl James. 2013.
\newblock \emph{Errors in language learning and use: Exploring error analysis}.
\newblock Routledge.

\bibitem[{Jiang et~al.(2023{\natexlab{a}})Jiang, Sablayrolles, Mensch, Bamford,
  Chaplot, Casas, Bressand, Lengyel, Lample, Saulnier
  et~al.}]{jiang2023mistral}
Albert~Q Jiang, Alexandre Sablayrolles, Arthur Mensch, Chris Bamford,
  Devendra~Singh Chaplot, Diego de~las Casas, Florian Bressand, Gianna Lengyel,
  Guillaume Lample, Lucile Saulnier, et~al. 2023{\natexlab{a}}.
\newblock Mistral 7b.
\newblock \emph{arXiv preprint arXiv:2310.06825}.

\bibitem[{Jiang et~al.(2023{\natexlab{b}})Jiang, Shi, Wang, Zhang, Zhang, Pan,
  Kim, and Li}]{jiang2023speech}
Ridong Jiang, Wei Shi, Bin Wang, Chen Zhang, Yan Zhang, Chunlei Pan, Jung~Jae
  Kim, and Haizhou Li. 2023{\natexlab{b}}.
\newblock Speech-aware multi-domain dialogue state generation with asr error
  correction modules.
\newblock In \emph{Proceedings of The Eleventh Dialog System Technology
  Challenge}, pages 105--112.

\bibitem[{Khot et~al.(2022)Khot, Trivedi, Finlayson, Fu, Richardson, Clark, and
  Sabharwal}]{khot2022decomposed}
Tushar Khot, Harsh Trivedi, Matthew Finlayson, Yao Fu, Kyle Richardson, Peter
  Clark, and Ashish Sabharwal. 2022.
\newblock Decomposed prompting: A modular approach for solving complex tasks.
\newblock \emph{arXiv preprint arXiv:2210.02406}.

\bibitem[{Kojima et~al.(2022)Kojima, Gu, Reid, Matsuo, and
  Iwasawa}]{kojima2022large}
Takeshi Kojima, Shixiang~Shane Gu, Machel Reid, Yutaka Matsuo, and Yusuke
  Iwasawa. 2022.
\newblock Large language models are zero-shot reasoners.
\newblock \emph{Advances in neural information processing systems},
  35:22199--22213.

\bibitem[{Leng et~al.(2021)Leng, Tan, Zhu, Xu, Luo, Liu, Qin, Li, Lin, and
  Liu}]{leng2021fastcorrect}
Yichong Leng, Xu~Tan, Linchen Zhu, Jin Xu, Renqian Luo, Linquan Liu, Tao Qin,
  Xiangyang Li, Edward Lin, and Tie-Yan Liu. 2021.
\newblock Fastcorrect: Fast error correction with edit alignment for automatic
  speech recognition.
\newblock \emph{Advances in Neural Information Processing Systems},
  34:21708--21719.

\bibitem[{Li et~al.(2018)Li, Ji, Du, Li, and Wang}]{li2018textbugger}
Jinfeng Li, Shouling Ji, Tianyu Du, Bo~Li, and Ting Wang. 2018.
\newblock Textbugger: Generating adversarial text against real-world
  applications.
\newblock \emph{arXiv preprint arXiv:1812.05271}.

\bibitem[{Lin et~al.(2024)Lin, Wang, Liu, and Chen}]{lin2024crossin}
Geyu Lin, Bin Wang, Zhengyuan Liu, and Nancy~F Chen. 2024.
\newblock Crossin: An efficient instruction tuning approach for cross-lingual
  knowledge alignment.
\newblock \emph{arXiv preprint arXiv:2404.11932}.

\bibitem[{Lovenia et~al.(2024)Lovenia, Mahendra, Akbar, Miranda, Santoso, Aco,
  Fadhilah, Mansurov, Imperial, Kampman et~al.}]{lovenia2024seacrowd}
Holy Lovenia, Rahmad Mahendra, Salsabil~Maulana Akbar, Lester James~V Miranda,
  Jennifer Santoso, Elyanah Aco, Akhdan Fadhilah, Jonibek Mansurov,
  Joseph~Marvin Imperial, Onno~P Kampman, et~al. 2024.
\newblock Seacrowd: A multilingual multimodal data hub and benchmark suite for
  southeast asian languages.
\newblock \emph{arXiv preprint arXiv:2406.10118}.

\bibitem[{Ma(2019)}]{ma2019nlpaug}
Edward Ma. 2019.
\newblock Nlp augmentation.
\newblock https://github.com/makcedward/nlpaug.

\bibitem[{Ma et~al.(2023)Ma, Qian, Manakul, Gales, and Knill}]{ma2023can}
Rao Ma, Mengjie Qian, Potsawee Manakul, Mark Gales, and Kate Knill. 2023.
\newblock Can generative large language models perform asr error correction?
\newblock \emph{arXiv preprint arXiv:2307.04172}.

\bibitem[{Mai and Carson-Berndsen(2024)}]{mai2024enhancing}
Long Mai and Julie Carson-Berndsen. 2024.
\newblock Enhancing conversation smoothness in language learning chatbots: An
  evaluation of gpt4 for asr error correction.
\newblock In \emph{ICASSP 2024-2024 IEEE International Conference on Acoustics,
  Speech and Signal Processing (ICASSP)}, pages 11001--11005. IEEE.

\bibitem[{Mani et~al.(2020)Mani, Palaskar, Meripo, Konam, and
  Metze}]{mani2020asr}
Anirudh Mani, Shruti Palaskar, Nimshi~Venkat Meripo, Sandeep Konam, and Florian
  Metze. 2020.
\newblock Asr error correction and domain adaptation using machine translation.
\newblock In \emph{ICASSP 2020-2020 IEEE International Conference on Acoustics,
  Speech and Signal Processing (ICASSP)}, pages 6344--6348. IEEE.

\bibitem[{Mesnard et~al.(2024)Mesnard, Hardin, Dadashi, Bhupatiraju, Sifre,
  Rivière, Kale, Love, Tafti, Hussenot, and et~al.}]{gemma}
Thomas Mesnard, Cassidy Hardin, Robert Dadashi, Surya Bhupatiraju, Laurent
  Sifre, Morgane Rivière, Mihir~Sanjay Kale, Juliette Love, Pouya Tafti,
  Léonard Hussenot, and et~al. 2024.
\newblock \href {https://doi.org/10.34740/KAGGLE/M/3301} {Gemma}.
\newblock \emph{Kaggle}.

\bibitem[{Napoles et~al.(2017)Napoles, Sakaguchi, and
  Tetreault}]{napoles-etal-2017-jfleg}
Courtney Napoles, Keisuke Sakaguchi, and Joel Tetreault. 2017.
\newblock \href {https://aclanthology.org/E17-2037} {{JFLEG}: A fluency corpus
  and benchmark for grammatical error correction}.
\newblock In \emph{Proceedings of the 15th Conference of the {E}uropean Chapter
  of the Association for Computational Linguistics: Volume 2, Short Papers},
  pages 229--234, Valencia, Spain. Association for Computational Linguistics.

\bibitem[{Nguyen et~al.(2021)Nguyen, Jatowt, Coustaty, and
  Doucet}]{nguyen2021survey}
Thi Tuyet~Hai Nguyen, Adam Jatowt, Mickael Coustaty, and Antoine Doucet. 2021.
\newblock Survey of post-ocr processing approaches.
\newblock \emph{ACM Computing Surveys (CSUR)}, 54(6):1--37.

\bibitem[{Panayotov et~al.(2015)Panayotov, Chen, Povey, and
  Khudanpur}]{panayotov2015librispeech}
Vassil Panayotov, Guoguo Chen, Daniel Povey, and Sanjeev Khudanpur. 2015.
\newblock Librispeech: an asr corpus based on public domain audio books.
\newblock In \emph{2015 IEEE international conference on acoustics, speech and
  signal processing (ICASSP)}, pages 5206--5210. IEEE.

\bibitem[{Prasad et~al.(2023)Prasad, Hase, Zhou, and
  Bansal}]{prasad-etal-2023-grips}
Archiki Prasad, Peter Hase, Xiang Zhou, and Mohit Bansal. 2023.
\newblock \href {https://doi.org/10.18653/v1/2023.eacl-main.277} {{G}r{IPS}:
  Gradient-free, edit-based instruction search for prompting large language
  models}.
\newblock In \emph{Proceedings of the 17th Conference of the European Chapter
  of the Association for Computational Linguistics}, pages 3845--3864,
  Dubrovnik, Croatia. Association for Computational Linguistics.

\bibitem[{Radford et~al.(2023)Radford, Kim, Xu, Brockman, McLeavey, and
  Sutskever}]{radford2023robust}
Alec Radford, Jong~Wook Kim, Tao Xu, Greg Brockman, Christine McLeavey, and
  Ilya Sutskever. 2023.
\newblock Robust speech recognition via large-scale weak supervision.
\newblock In \emph{International Conference on Machine Learning}, pages
  28492--28518. PMLR.

\bibitem[{Sclar et~al.(2023)Sclar, Choi, Tsvetkov, and
  Suhr}]{sclar2023quantifying}
Melanie Sclar, Yejin Choi, Yulia Tsvetkov, and Alane Suhr. 2023.
\newblock Quantifying language models' sensitivity to spurious features in
  prompt design or: How i learned to start worrying about prompt formatting.
\newblock \emph{arXiv preprint arXiv:2310.11324}.

\bibitem[{Soper et~al.(2021)Soper, Fujimoto, and Yu}]{soper-etal-2021-bart}
Elizabeth Soper, Stanley Fujimoto, and Yen-Yun Yu. 2021.
\newblock \href {https://doi.org/10.18653/v1/2021.wnut-1.31} {{BART} for
  post-correction of {OCR} newspaper text}.
\newblock In \emph{Proceedings of the Seventh Workshop on Noisy User-generated
  Text (W-NUT 2021)}, pages 284--290, Online. Association for Computational
  Linguistics.

\bibitem[{Stahlberg and Kumar(2021)}]{stahlberg-kumar-2021-synthetic}
Felix Stahlberg and Shankar Kumar. 2021.
\newblock \href {https://aclanthology.org/2021.bea-1.4} {Synthetic data
  generation for grammatical error correction with tagged corruption models}.
\newblock In \emph{Proceedings of the 16th Workshop on Innovative Use of NLP
  for Building Educational Applications}, pages 37--47, Online. Association for
  Computational Linguistics.

\bibitem[{Tang et~al.(2024)Tang, Yu, Sun, Chen, Tan, Li, Lu, MA, and
  Zhang}]{tang2024salmonn}
Changli Tang, Wenyi Yu, Guangzhi Sun, Xianzhao Chen, Tian Tan, Wei Li, Lu~Lu,
  Zejun MA, and Chao Zhang. 2024.
\newblock \href {https://openreview.net/forum?id=14rn7HpKVk} {{SALMONN}:
  Towards generic hearing abilities for large language models}.
\newblock In \emph{The Twelfth International Conference on Learning
  Representations}.

\bibitem[{Team et~al.(2023)Team, Anil, Borgeaud, Wu, Alayrac, Yu, Soricut,
  Schalkwyk, Dai, Hauth et~al.}]{team2023gemini}
Gemini Team, Rohan Anil, Sebastian Borgeaud, Yonghui Wu, Jean-Baptiste Alayrac,
  Jiahui Yu, Radu Soricut, Johan Schalkwyk, Andrew~M Dai, Anja Hauth, et~al.
  2023.
\newblock Gemini: a family of highly capable multimodal models.
\newblock \emph{arXiv preprint arXiv:2312.11805}.

\bibitem[{Tong and Evans(1996)}]{tong-evans-1996-statistical}
Xiang Tong and David~A. Evans. 1996.
\newblock \href {https://aclanthology.org/W96-0108} {A statistical approach to
  automatic {OCR} error correction in context}.
\newblock In \emph{Fourth Workshop on Very Large Corpora}, Herstmonceux Castle,
  Sussex, UK. Association for Computational Linguistics.

\bibitem[{Touvron et~al.(2023)Touvron, Martin, Stone, Albert, Almahairi,
  Babaei, Bashlykov, Batra, Bhargava, Bhosale et~al.}]{touvron2023llama}
Hugo Touvron, Louis Martin, Kevin Stone, Peter Albert, Amjad Almahairi, Yasmine
  Babaei, Nikolay Bashlykov, Soumya Batra, Prajjwal Bhargava, Shruti Bhosale,
  et~al. 2023.
\newblock Llama 2: Open foundation and fine-tuned chat models.
\newblock \emph{arXiv preprint arXiv:2307.09288}.

\bibitem[{Tunstall et~al.(2023)Tunstall, Beeching, Lambert, Rajani, Rasul,
  Belkada, Huang, von Werra, Fourrier, Habib et~al.}]{tunstall2023zephyr}
Lewis Tunstall, Edward Beeching, Nathan Lambert, Nazneen Rajani, Kashif Rasul,
  Younes Belkada, Shengyi Huang, Leandro von Werra, Cl{\'e}mentine Fourrier,
  Nathan Habib, et~al. 2023.
\newblock Zephyr: Direct distillation of lm alignment.
\newblock \emph{arXiv preprint arXiv:2310.16944}.

\bibitem[{Wang et~al.(2024{\natexlab{a}})Wang, Lin, Liu, Wei, and
  Chen}]{wang-etal-2024-craft}
Bin Wang, Geyu Lin, Zhengyuan Liu, Chengwei Wei, and Nancy Chen.
  2024{\natexlab{a}}.
\newblock \href {https://doi.org/10.18653/v1/2024.c3nlp-1.4} {{CRAFT}:
  Extracting and tuning cultural instructions from the wild}.
\newblock In \emph{Proceedings of the 2nd Workshop on Cross-Cultural
  Considerations in NLP}, pages 42--47, Bangkok, Thailand. Association for
  Computational Linguistics.

\bibitem[{Wang et~al.(2024{\natexlab{b}})Wang, Liu, Huang, Jiao, Ding, Aw, and
  Chen}]{wang-etal-2024-seaeval}
Bin Wang, Zhengyuan Liu, Xin Huang, Fangkai Jiao, Yang Ding, AiTi Aw, and Nancy
  Chen. 2024{\natexlab{b}}.
\newblock \href {https://doi.org/10.18653/v1/2024.naacl-long.22} {{S}ea{E}val
  for multilingual foundation models: From cross-lingual alignment to cultural
  reasoning}.
\newblock In \emph{Proceedings of the 2024 Conference of the North American
  Chapter of the Association for Computational Linguistics: Human Language
  Technologies (Volume 1: Long Papers)}, pages 370--390, Mexico City, Mexico.
  Association for Computational Linguistics.

\bibitem[{Wang et~al.(2024{\natexlab{c}})Wang, Zou, Lin, Sun, Liu, Zhang, Liu,
  Aw, and Chen}]{wang2024audiobench}
Bin Wang, Xunlong Zou, Geyu Lin, Shuo Sun, Zhuohan Liu, Wenyu Zhang, Zhengyuan
  Liu, AiTi Aw, and Nancy~F Chen. 2024{\natexlab{c}}.
\newblock Audiobench: A universal benchmark for audio large language models.
\newblock \emph{arXiv preprint arXiv:2406.16020}.

\bibitem[{Wei et~al.(2023)Wei, Wang, Wang, and Kuo}]{wei2023overview}
Chengwei Wei, Yun-Cheng Wang, Bin Wang, and C-C~Jay Kuo. 2023.
\newblock An overview on language models: Recent developments and outlook.
\newblock \emph{arXiv preprint arXiv:2303.05759}.

\bibitem[{Xue et~al.(2022)Xue, Barua, Constant, Al-Rfou, Narang, Kale, Roberts,
  and Raffel}]{xue2022byt5}
Linting Xue, Aditya Barua, Noah Constant, Rami Al-Rfou, Sharan Narang, Mihir
  Kale, Adam Roberts, and Colin Raffel. 2022.
\newblock Byt5: Towards a token-free future with pre-trained byte-to-byte
  models.
\newblock \emph{Transactions of the Association for Computational Linguistics},
  10:291--306.

\bibitem[{Yuan and Briscoe(2016)}]{yuan-briscoe-2016-grammatical}
Zheng Yuan and Ted Briscoe. 2016.
\newblock \href {https://doi.org/10.18653/v1/N16-1042} {Grammatical error
  correction using neural machine translation}.
\newblock In \emph{Proceedings of the 2016 Conference of the North {A}merican
  Chapter of the Association for Computational Linguistics: Human Language
  Technologies}, pages 380--386, San Diego, California. Association for
  Computational Linguistics.

\bibitem[{Zhang et~al.(2024{\natexlab{a}})Zhang, Zeng, Wang, and
  Lu}]{zhang2024tinyllama}
Peiyuan Zhang, Guangtao Zeng, Tianduo Wang, and Wei Lu. 2024{\natexlab{a}}.
\newblock \href {http://arxiv.org/abs/2401.02385} {Tinyllama: An open-source
  small language model}.

\bibitem[{Zhang et~al.(2020)Zhang, Huang, Liu, and
  Li}]{zhang-etal-2020-spelling}
Shaohua Zhang, Haoran Huang, Jicong Liu, and Hang Li. 2020.
\newblock \href {https://doi.org/10.18653/v1/2020.acl-main.82} {Spelling error
  correction with soft-masked {BERT}}.
\newblock In \emph{Proceedings of the 58th Annual Meeting of the Association
  for Computational Linguistics}, pages 882--890, Online. Association for
  Computational Linguistics.

\bibitem[{Zhang et~al.(2024{\natexlab{b}})Zhang, Sun, Wang, Zou, Liu, He, Lin,
  Chen, and Aw}]{zhang2024mowe}
Wenyu Zhang, Shuo Sun, Bin Wang, Xunlong Zou, Zhuohan Liu, Yingxu He, Geyu Lin,
  Nancy~F Chen, and Ai~Ti Aw. 2024{\natexlab{b}}.
\newblock Mowe-audio: Multitask audiollms with mixture of weak encoders.
\newblock \emph{arXiv preprint arXiv:2409.06635}.

\bibitem[{Zhou et~al.(2023)Zhou, Muresanu, Han, Paster, Pitis, Chan, and
  Ba}]{zhou2022large}
Yongchao Zhou, Andrei~Ioan Muresanu, Ziwen Han, Keiran Paster, Silviu Pitis,
  Harris Chan, and Jimmy Ba. 2023.
\newblock Large language models are human-level prompt engineers.
\newblock \emph{ICLR}.

\bibitem[{Zhu et~al.(2023)Zhu, Wang, Zhou, Wang, Chen, Wang, Yang, Ye, Gong,
  Zhang et~al.}]{zhu2023promptbench}
Kaijie Zhu, Jindong Wang, Jiaheng Zhou, Zichen Wang, Hao Chen, Yidong Wang,
  Linyi Yang, Wei Ye, Neil~Zhenqiang Gong, Yue Zhang, et~al. 2023.
\newblock Promptbench: Towards evaluating the robustness of large language
  models on adversarial prompts.
\newblock \emph{arXiv preprint arXiv:2306.04528}.

\end{thebibliography}
\appendix

\section{Instruction Templates}
\label{sec_app:prompt}

    %%%%%%%%%%%%%%%%%%%%%%%%%%%%%%%%%%%%%%%%%%%%%%%
    \begin{table*}[t]
        \centering
        \begin{adjustbox}{width=1.00\textwidth,center}
        \begin{tabular}{ | c | c | }
        \toprule
        \textbf{Task} & \textbf{Prompt Template} \\ 
        \hline
        \begin{tabular}{@{}c@{}}
        \textbf{Noise Simulation} \\ 
        \end{tabular} 
        & 
        \begin{tabular}{@{}c@{}} 
        \texttt{[Uttrance\_1] Please help me generate errors in the sentence:} \\ 
        \texttt{<< \$\{Instruction\} >> [/Uttrance\_1]} 
        \end{tabular} 
        \\\hline

        \begin{tabular}{@{}c@{}}
        \textbf{Error Corrections} \\
        (\emph{ChatGPT-3.5})
        \end{tabular} 
        & 
        \begin{tabular}{@{}c@{}} 
        \texttt{[Uttrance\_1] You are an error correction assistant. Do not output} \\ 
        \texttt{additional explanations besides the corrected instruction. [/Uttrance\_1]} \\
        \texttt{[Uttrance\_2] Please help me correct the instruction if it contains any error.} \\
        \texttt{Instruction: \$\{Instruction\}. Corrected Instruction: [/Uttrance\_2]} 
        \end{tabular} 
        \\ \hline

        \begin{tabular}{@{}c@{}}
        \textbf{Error Corrections} \\
        (\emph{Mistral-7B-Instruct-v0.2})
        \end{tabular} 
        & 
        \begin{tabular}{@{}c@{}} 
        \texttt{[Uttrance\_1] You are a chatbot who always responds with corrected instructions.} \\ 
        \texttt{[/Uttrance\_1] [Uttrance\_2] No problem! I will just correct the errors in the} \\
        \texttt{content without any other output. Let's get started! [/Uttrance\_2] [Uttrance\_3]} \\
        \texttt{Please help me correct possible errors in the instruction. Do not output anything} \\
        \texttt{else. Instruction: \$\{Instruction\} Corrected Instruction: [/Uttrance\_3]}
        \end{tabular} 
        \\ \hline

        \begin{tabular}{@{}c@{}}
        \textbf{Error Corrections} \\
        (\emph{Llama-2-7B-Chat})
        \end{tabular} 
        & 
        \begin{tabular}{@{}c@{}} 
        \texttt{[Uttrance\_1] You are a chatbot who always responds with corrected instructions.} \\ 
        \texttt{ [/Uttrance\_1] [Uttrance\_2] No problem! I will just correct the errors in } \\
        \texttt{the content and output the corrected content without any other outputs.} \\
        \texttt{[/Uttrance\_2] [Uttrance\_3] Please help me correct possible errors in} \\
        \texttt{the instruction. Do not output anything else. Instruction: \$\{Instruction\}  } \\
        \texttt{Corrected Instruction: [/Uttrance\_3]}

        \end{tabular} 
        \\ 
        
        \bottomrule
        \end{tabular}
        \end{adjustbox}
        \caption{Templates for simulating noise and correcting errors. The dialogue template adheres to the format specified by the respective models.}
        \label{tab:prompt-templates}
    \end{table*}
    %%%%%%%%%%%%%%%%%%%%%%%%%%%%%%%%%%%%%%%%%%%%%%%

\end{document}